\documentclass[10pt,journal]{IEEEtran}

\usepackage[caption=false,font=footnotesize]{subfig}

\usepackage[hidelinks]{hyperref}
\usepackage{amssymb,amsfonts}
\usepackage{mathtools}
\usepackage{graphicx}
\usepackage{xpatch}
\usepackage{dblfloatfix}    

\usepackage[caption=false, font=footnotesize]{subfig}

\usepackage[utf8]{inputenc}

\usepackage{sansmath}
\usepackage[defaultmathsizes, italic]{mathastext}

\usepackage{ifdraft}
\usepackage{etoolbox}
\usepackage{mfirstuc}
\usepackage{acro}
\acsetup{single=true}

\NewAcroTemplate{long-short-nop}{\acrowrite{long} \acrowrite{short}}

\DeclareAcronym{gpu}{short = GPU, long = general-purpose graphics processing unit, first-style=short, single-style=short}
\DeclareAcronym{cpu}{short = CPU, long = central processing unit, first-style=short, single-style=short}

\DeclareAcronym{vth}{short = V\kern-0.1em\textsubscript{th}, long = threshold voltage, first-style=long-short-nop}
\DeclareAcronym{dvt}{short = $\Delta$\kern-0.15em\acs*{vth}, long = $\Delta$\acs*{vth}, first-style=short, pdfstring=delta Vth}
\DeclareAcronym{vg}{short = V\textsubscript{G}, long = gate voltage, first-style=long-short-nop}
\DeclareAcronym{vdd}{short = V\textsubscript{DD}, long = supply voltage, first-style=long-short-nop}
\DeclareAcronym{hfo}{short = HfO\textsubscript{2}, long = Hafnium Oxide }
\DeclareAcronym{highk}{short = high\nobreakdash-$\kappa$, long = high\nobreakdash-$\kappa$, first-style=short}
\DeclareAcronym{bti}{short = NBTI, long = negative bias temperature instability, single-style=long-short}
\DeclareAcronym{hcd}{short = HCD, long = hot carrier degradation, single-style=long-short}
\DeclareAcronym{bat}{short=BAT, long=BTI Analysis Tool}

\DeclareAcronym{hd}{short=HD, long=hyperdimensional, short-indefinite=an, long-indefinite=a}
\DeclareAcronym{hdc}{short=HDC, long=hyperdimensional computing, short-indefinite=an}
\DeclareAcronym{hdv}{short=hypervector, long=hypervector, first-style=short}
\DeclareAcronym{im}{short=IM, long=item memory}

\DeclareAcronym{ml}{short = ML, long = machine learning, short-indefinite=an}
\DeclareAcronym{dnn}{short = DNN, long = deep neural network}
\DeclareAcronym{rnn}{short = RNN, long = recurrent neural network}
\DeclareAcronym{lstm}{short = LSTM, long = long short-term memory, short-indefinite=an}
\DeclareAcronym{svm}{short = SVM, long = support vector machine, short-indefinite=an, long-indefinite=a}
\DeclareAcronym{mlp}{short = MLP, long = multilayer perceptron, short-indefinite=an}
\DeclareAcronym{pdk}{short = PDK, long = process design kit}

\DeclareAcronym{wf}{short = waveform, long = voltage waveform, first-style=long}
\DeclareAcronym{trace}{short = trace, long = degradation trace, first-style=long}
\DeclareAcronym{re}{short = $RE$, long = relative error}
\DeclareAcronym{eol}{short=EoL, long = end of lifetime}

\usepackage[per-mode=symbol,group-separator={,},detect-all, list-final-separator={, and }]{siunitx}

\usepackage{pgfplots}
\usepackage{circledsteps}
\tikzset{/csteps/inner color=white, /csteps/fill color=black}

\tikzset{
    font={\fontsize{8pt}{9}\selectfont},
    arrow/.style={-latex}
    }

\usepackage{uni-stuttgart-colors}

\usepackage[noabbrev,capitalise]{cleveref}
\Crefname{subsection}{Section}{Sections}
\Crefname{subsubsection}{Section}{Sections}
\Crefname{paragraph}{Section}{Sections}
\Crefname{figure}{Fig.}{Fig.}
\Crefname{table}{Tab.}{Tab.}

\usepackage{microtype}
\usepackage[normalem]{ulem}
\usepackage{tabularx}
\usepackage{booktabs}
\usepackage{multirow}
\usepackage{makecell}
\usepackage{textcomp}
\usepackage[bibstyle=ieee, citestyle=numeric-comp, maxbibnames=5, minbibnames=5, maxcitenames=1, mincitenames=1, doi=false, isbn=false, url=false, 
backend=biber]{biblatex}
\DeclareFieldFormat{journaltitle}{#1\isdot}
\DeclareFieldFormat[book]{title}{{#1\isdot}}

\AtEveryBibitem{
 \clearname{editor}
 \clearfield{month}
  \clearfield{extra}
}
\DeclareSourcemap{
  \maps[datatype=bibtex]{
    \map{
      \step[fieldset=note, null]
    }
  }
}
\DefineBibliographyStrings{english}{%
    andothers = {et al\adddot}
}

\usepackage{enumitem}
\usepackage{balance}

\makeatletter
\g@addto@macro\@floatboxreset{\centering}
\makeatother

\title{Modeling and Predicting Transistor Aging under Workload Dependency using Machine Learning
}
\begin{document}

\author{Paul~R.~Genssler,~\IEEEmembership{Member,~IEEE},
        Hamza~E.~Barkam,~\IEEEmembership{Member,~IEEE},
        Karthik~Pandaram,
        Mohsen~Imani,~\IEEEmembership{Member,~IEEE} and 
        Hussam~Amrouch,~\IEEEmembership{Member,~IEEE}
	\thanks{%
		Paul R. Genssler, Karthik Pandaram, and Hussam Amrouch are with the Chair of Semiconductor Test and Reliability (STAR), University of Stuttgart, Stuttgart 70569, Germany. E-mail: \{genssler, amrouch\}@iti.uni-stuttgart.de, st174730@stud.uni-stuttgart.de. Hamza E. Barkam and Mohsen Imani are with the Bio-Inspired Architecture and Systems (BIASLab), UC Irvine, California, USA. E-mail: \{herrahmo, m.imani\}@uci.edu.
	}%
}

\IEEEoverridecommandlockouts
\IEEEpubid{\begin{minipage}{1.1\textwidth}\ \\[25pt]This work has been submitted to the IEEE for possible publication. Copyright may be transferred without notice, after which this version may no longer be accessible.\end{minipage}}

\maketitle

\begin{abstract}
The pivotal issue of reliability is one of colossal concern for circuit designers. The driving force is transistor aging, dependent on operating voltage and workload. At the design time, it is difficult to estimate close-to-the-edge guardbands that keep aging effects during the lifetime at bay. This is because the foundry does not share its calibrated physics-based models, comprised of highly confidential technology and material parameters. However, the unmonitored yet necessary overestimation of degradation amounts to a performance decline, which could be preventable. Furthermore, these physics-based models are exceptionally computationally complex. The costs of modeling millions of individual transistors at design time can be evidently exorbitant. We propose the revolutionizing prospect of a machine learning model trained to replicate the physics-based model, such that no confidential parameters are disclosed. This effectual workaround is fully accessible to circuit designers for the purposes of design optimization. We demonstrate the models' ability to generalize by training on data from one circuit and applying it successfully to a benchmark circuit. The mean relative error is as low as \SI{1.7}{\percent}, with a speedup of up to 20X. Circuit designers, for the first time ever, will have ease of access to a high-precision aging model, which is paramount for efficient designs. This work is a promising step in the direction of bridging the wide gulf between the foundry and circuit designers.

\end{abstract}


\section{Introduction}\label{sec:introduction}

\acresetall
\IEEEPARstart{R}{eliability} is a major concern in today's circuits.
As CMOS scaling reaches the atomic level, the impact of degradation effects on the reliability becomes stronger \cite{comprehensivedeviceproduct2018huang}.
Aging is the most dominating effect and changes the transistor's properties like the \ac{vth}.
Consequently, it can cause permanent failures in a circuit.
Even before such failures, aging indirectly impacts the circuit's timing and hinders performance improvements.
The \ac{bti} aging mechanism is responsible for the highest degradation \cite{modelingnbtiusing2020mahapatra}.
During regular transistor operation, Si-H bonds at the Si-SiO\textsubscript{2} interface might be broken and annealed.
Additionally, charges are captured and emitted in the oxide vacancies at the interface layer.
Over time, these defects accumulate and manifest themselves as a shift in \ac{vth}, referred to as \ac{dvt}.
The induced increase in the propagation delay of the logic gates can cause timing violations.

To prevent such timing violations and ensure the circuit performs as specified during its entire projected lifetime, timing guardbands are added during the design phase.
Such additional slack compensates for the reduced switching speed of aged transistors.
The design challenge is to balance such guardbands between too pessimistic, reducing the circuit's performance, and too optimistic, increasing the risk of premature failures.
To find an optimal guardband (i.e. small, yet sufficient), the aging-induced \ac{dvt} has to be accurately estimated. 
Aging models are required to abstract the underlying physical behaviors, take technology parameters, stress patterns, and voltages into account, and predict the evolution of \ac{dvt} over time.
Only with such models can designers make informed and proper decisions on the guardband of every transistor.

\begin{figure}
    \includegraphics{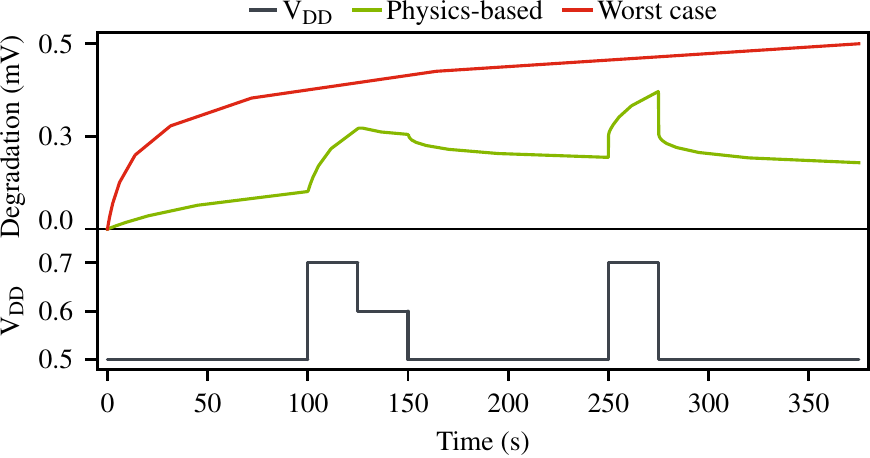}
    \caption{Worst-case models are typically employed in the industry. For transistor aging, they assume constant stress and thus the highest possible degradation (red). Physics-based models are far more accurate because they take the input waveform and recovery effects into account.}
    \label{fig:aging motivation}
\end{figure}

\begin{figure*}[t]
    \includegraphics{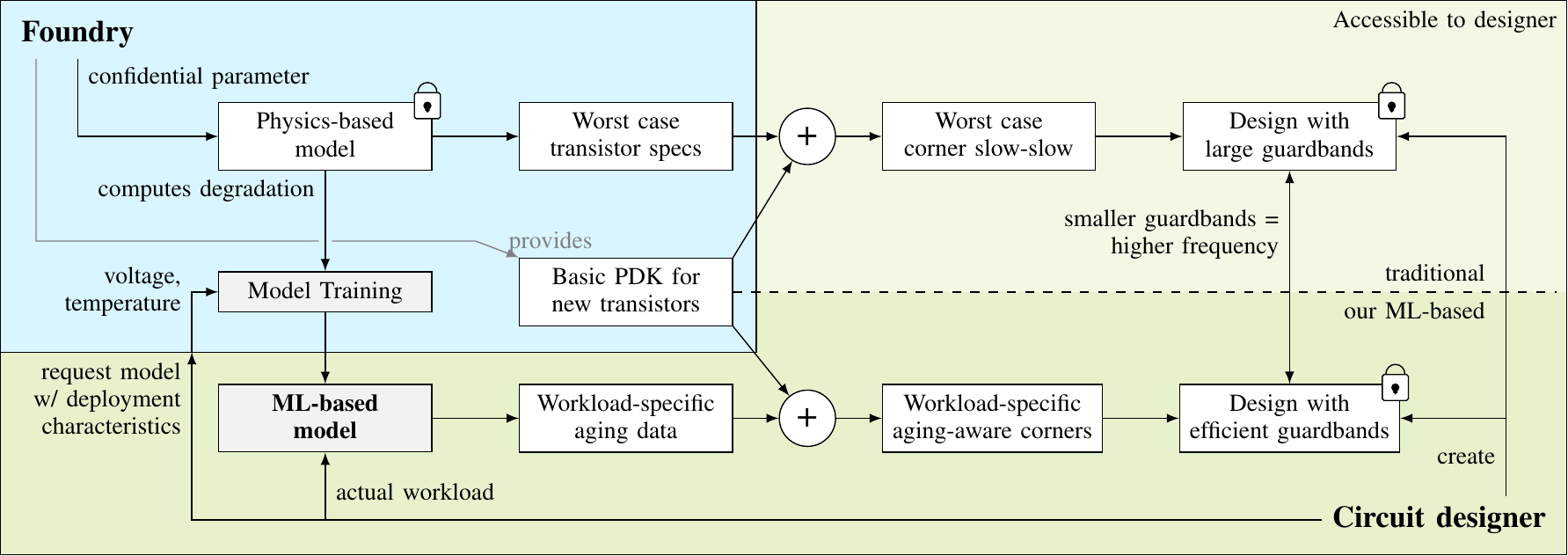}
    \caption{Typically, circuit designers do not have access to accurate physics-based aging models to estimate efficient (i.e., small, yet sufficient) guardbands. \Iacl*{ml}-based aging model are free sensitive material and process parameters of the foundry and can thus be shared with designers. Now, circuit designer can create workload-specific aging data for efficient guardbands.}
    \label{fig:overview}
    \vspace{-0.5ex}
\end{figure*}

Physics-based aging models capture the \textit{dynamics} of the fundamental physical behavior and chemical reactions inside the transistors.
Complex differential equations take the material and technology dependent parameters into account. 
This makes the model capable of capturing recovery effects, where \ac{vth} is indeed reduced as shown in \cref{fig:aging motivation}.
During low-stress phases, the defects are partially healed and \ac{vth} recovers \cite{revolvingreferenceodometer2015satapathy}.
The \acf{vdd} is dynamic, creating such phases, changes over time, and is typically defined through the workload of the circuit.
To capture these voltage dynamics, an aging model has to process such \iacl*{wf}.
Worst-case aging models are not capable of this.
They are created by fitting measurements of constant voltage stress on a transistor.
Hence, they cannot model the physics of voltage dynamics and recovery effects.
To process \iacl*{wf}, the highest voltage is applied for the whole duration of the \ac{wf}.
Consequently, they overestimate the impact of aging significantly.
Today's high-end devices are operating at the technological limits and cannot afford the unnecessary performance penalties mandated by such pessimistic predictions, an ideal aging model has to be as precise as possible.
While physics-based models achieve such high accuracy, they require parameters specific to the manufacturing process to compute the degradation.
Such parameters are a valuable secret of the foundry because they reveal details about their technology through material-dependent parameters.
The foundry instead provides \iac{pdk} covering various corner cases including the worst case (i.e., the slow-slow corner).
In summary, designers have limited options to optimize their circuit, which reduces performance and increases costs.
An ``ideal'' aging model should therefore not expose any confidential information about the underlying technology.
At the same time, it should still provide accurate estimations, including recovery.

The foundry only guarantees the slow-slow corner leading to very pessimistic guardbands and hence efficiency losses.
With the risk of failure on the designers' side, this pessimism might be reduced.
Alternatively, the degradation can be measured during post-silicon validation.
However, at this stage, the design is almost complete making changes costly.
With an aging model, the impact of the circuit's workloads and voltages on \ac{vth} can be predicted early in the design phase.
Starting with the much faster typical-typical corner, an appropriate guardband is added.
An ideal aging model is thus available to the designers during design time and allows them to predict the degradation for each individual transistor.
During runtime, the remaining guardband can be treated as a resource like remaining battery power.
Resource management schemes require a long-term aging model to optimize over the whole lifetime.
Physics-based models are not an option, because of their confidential parameters and their high computational complexity.
An ideal aging model has a low computational cost to be employed for millions of transistors during design time.
At runtime, it provides predictions as a low-overhead background task in the operating system.

\textbf{Our Main Contributions}\\
Designers require an \textit{accurate and fast} transistor aging model to optimize the performance of their circuit designs depending on the potential workload.
Further, simulating millions and billions of transistors is time consuming necessitating a fast aging model.
Physics-based models are slow and confidential, i.e., not accessible to designers.
Therefore, we propose to employ \ac{ml} to model transistor aging.
As shown in \cref{fig:overview}, the foundry employs its confidential physics-based models to train \iac{ml}-based model.
Such a model is fast and does not reveal the technology and material parameters. 
Hence, it can be provided to the circuit designers.
They employ the model in conjunction with their workloads to generate their workload-specific, aging-aware \ac{pdk}.
With this \ac{pdk}, guardbands can be reduced increasing performance.

In this paper, we investigate for the first time how physics-based models can be abstracted through \ac{ml} methods. 
\ac{ml} algorithms like \ac{dnn} or \ac{lstm} have a high computational complexity but can achieve in high accuracy in many applications.
As a less computational-intense algorithm, lightweight brain-inspired \ac{ml} methods have attracted the interest of the community in recent years.
Brain-inspired \ac{hdc} does not utilize networks of neurons but is built around large randomly-generated \acp{hdv} \cite{hyperdimensionalcomputingintroduction2009kanerva}.
The accurate yet complex equations of physics-based models have to be replaced by a trained \ac{ml} model. 
To this end, we investigate two challenges. 
First, the capability to constructed \iac{dvt} trace from a voltage activity waveform. 
Such traces and waveforms are typically in the range of nanoseconds to minutes and model short-term aging \cite{van2016designing}. 
Second, predict only the last degradation \ac{dvt} value for a single transistor based on a given short voltage activity waveform. 
This prediction is essential for an extrapolation to ten years until the \ac{eol} of the device. 
We investigate the accuracy of the ML models not only on their prediction of this \ac{dvt} value.
We also employ the predicted \ac{dvt} further to extrapolate the circuit delay after ten years and compare the impact on the delay. 
The performance of the models is evaluated by training on the transistors of standard cells and an 8-bit adder. 
The test set are the transistors of a 32-bit MAC unit with which we also evaluate the prediction of the delay after 10 years.

\section{Related Work and Background}
Transistor aging has been studied for many years and the impact is well understood.
This sections aims at summarizing this research briefly.

\startcstep
\begin{figure*}[b]
    \centering
    \includegraphics{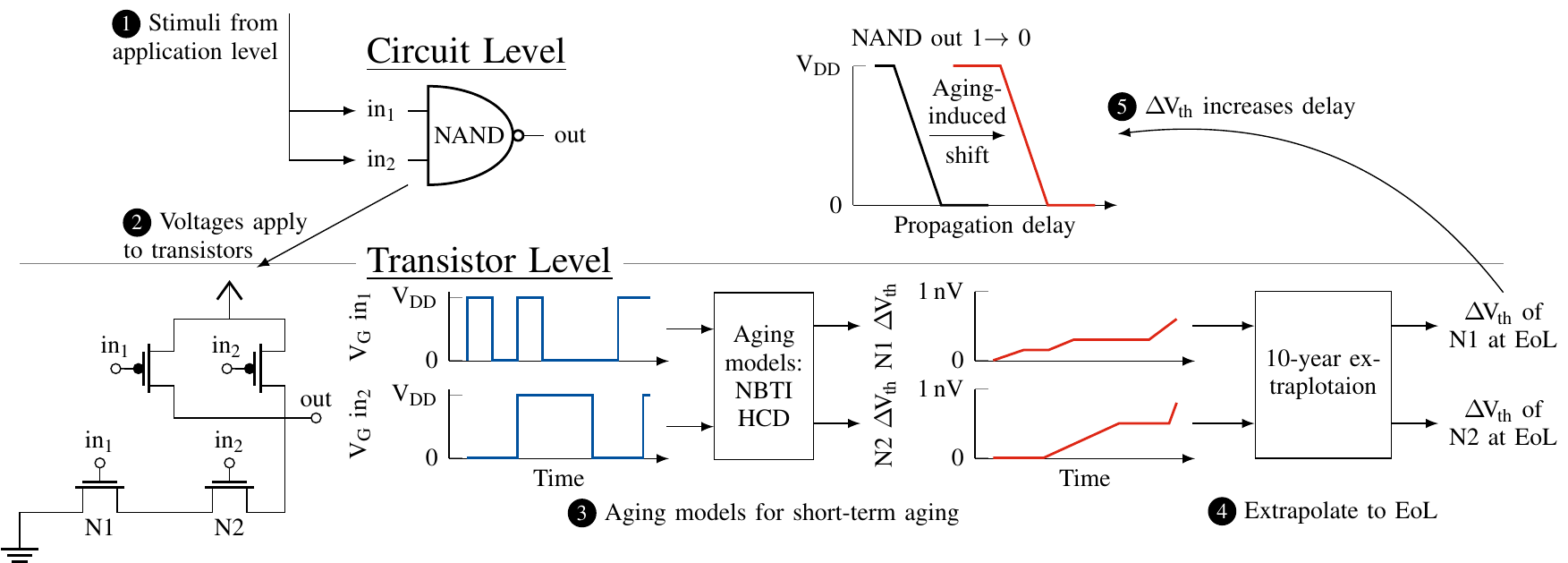}
    \caption{In the experimental setup, stimuli are applied at circuit level \Circled{1} and voltage waveforms for each transistor extracted  \Circled{2}. Those are passed to the aging models \Circled{3} to generate the ground truth for the training of the \acl*{ml} models. Then, their prediction is extrapolated to the \acf*{eol} \Circled{4}. Finally, the degradation is applied again at circuit level for efficient guardband estimation \Circled{5}.}
    \label{fig:outline}
\end{figure*}

\subsection{Transistor Aging Models}

Since manufacturing technology has moved past \SI{45}{\nano\meter}, new materials had to be used \cite{transistoraging2011keane}.
\Ac{hfo} is used as a \ac{highk} dielectric and replaced the traditional silicon dioxide.
A drawback of \ac{hfo} is its higher number of pre-existing defects in the material itself, making it more susceptible to degradation and thus less reliable.
Hence, transistor aging has become a major consideration in modern circuits.

In this work, we focus on \ac{bti} as the primary aging mechanism \cite{modelingnbtiusing2020mahapatra}.
Note that our method can be applied analogously to other aging mechanisms like \ac{hcd}.
\Ac{bti} aging occurs when the pMOS transistor is turned on.
During the on-time, two effects come into play.
First, positively charged holes are trapped inside the \ac{hfo} dielectric.
This increases the \ac{vth} of the transistor.
If the stress is reduced, i.e., the voltage lowered or the transistor completely turned off, then the holes can be removed and the initial \ac{vth} can be recovered over time.
Due to the second effect, new traps are generated in the interface material.
If the transistor is turned on, these traps are positively charged increasing the \ac{vth}.
Similar to the first effect, some of these traps may be deactivated once the stress is reduced or removed partially restoring \ac{vth}.
In both cases \ac{dvt} is dictated by the applied voltage.

Most models (especially analytical models) consider recovery only at \SI{0}{\volt}.
However, measurements have proven that even a reduction in the voltage starts the recovery \cite{revolvingreferenceodometer2015satapathy}.
The phenomenon is demonstrated in \cref{fig:aging motivation}, in which a physics-based \ac{bti} model is employed to calculate the transient trap occupancy, among others \cite{modelingnbtiusing2020mahapatra}.
Hence, it is indispensable to consider the dynamics of different voltage levels when modeling aging \cite{agingawarevoltagescaling2016vansanten}.

\Ac{ml}-based methods to model and predict the impact of aging have been investigated at different levels of the stack.
At the system level, reinforcement learning-based methods have been used to schedule threads on a multi-core \ac{cpu} to reduce aging \cite{reinforcementlearningbasedinter2014das}.
At the circuit level, the increase in path delay due to an increased \ac{dvt} has been modeled with multivariate adaptive regression splines and compared against \ac{svm} and \ac{rnn} \cite{realtimepredictionic2019huang}.
Their model takes changing operation conditions, like different voltages, into account.
At the gate level, the generation of reliability-aware cell libraries through \ac{ml} has been demonstrated \cite{machinelearningonthefly2021klemme}.
In \cite{machinelearningtransistor2021chatterjee}, at device level, a single transistor is subjected to constant voltage stress and the \ac{vth} curve is fitted with a regression model.
In this work, we are the first to explore the applicability of \ac{ml} methods at the device and physics level.
In contrast to \cite{machinelearningtransistor2021chatterjee}, we include voltage dynamics and recovery effects.
Further, the input to our model is not a single fixed voltage or a statistical assumption of on/off times, but a trace representing workloads and operating conditions for an individual transistor. 

\subsection{Machine-learning Methods}

As for our predictive models, we used different strategies and analyzed what were the trade-offs between each one of them. The \ac{mlp} model is one of the simplest neural network models and this practicality has caused its increase on popularity. On the other hand, \ac{ml} focused on the maximization (support) of separating the margin between classes (vector), also called \ac{svm} learning, is a powerful classification tool that has been used widely on many applications and achieved great results.

\Acp{rnn} are frequently used in application involving sequential data, which fits the temporal nature of aging. 
However, \acp{rnn} frequently fail to learn the important information from the input data involving learning long-term dependencies. 
By introducing gate functions into the cell structure, the \ac{lstm} is able to handle the problem of long-term dependencies well \cite{10.1162/neco.1997.9.8.1735}. 
Since its introduction, almost all the results based on \acp{rnn} have been achieved by \acp{lstm}.
The many applications include machine translation, time series prediction, natural language processing, and Computer Vision among others \cite{Priatama2022-jk}.
Because of the influence of previous voltages on aging, \ac{lstm}'s ability to successfully train on data with long-term temporal dependencies makes it natural choice for this application \cite{NIPS1996_a4d2f0d2}.

\subsection{Brain-Inspired Hyperdimensional Computing}

Brain-inspired \ac{hdc} is a lightweight alternative to traditional \ac{ml} approaches.
It is a rapidly emerging concept that has been successfully applied to voice recognition \cite{voicehdhyperdimensionalcomputing2017imani}, and hand gesture identification \cite{emggesturerecognition2018moin}, seizures detected \cite{ensemblehyperdimensionalclassifiers2020burrello}, image classification \cite{dynamichyperdimensionalcomputing2020chuang}, pattern recognition for wafer defect maps \cite{braininspiredcomputingwafer2021genssler}, circuit reliability estimation \cite{aspdac22_SS, sramTC2022}, and others.
Implementations range from low-power embedded devices \cite{revisitinghyperdimensionallearning2021imani} to high-power \acp{gpu} \cite{onlinehdrobustefficient2021hernandez-cane}.
\Ac{hdc} is based on the concept of \acp{hdv}, vectors with thousands of dimensions.
The \acp{hdv} can consist of simple bits, integers, real numbers, or other symbols.

\Acp{hdv} representing real-world values (e.g., \SI{0.7}{\volt}) are generated once and stored in the \acl{im}.
If the same value has to be mapped into \acl{hd} space again, the previously generated item \ac{hdv} is retrieved from the \acl{im}.
Due to the high dimension, it is very likely that two randomly-generated \acp{hdv} are orthogonal to each other. 
For binary \acp{hdv}, this similarity metric is computed with the Hamming distance, for integer-based \acp{hdv} using the cosine similarity.
 
Multiple item \acp{hdv} are combined into a class \ac{hdv} through the basic operations of bundling and binding \cite{hyperdimensionalcomputingintroduction2009kanerva}.
This process is also called encoding.
A voltage waveform is encoded into a single \ac{hdv} which then represents said waveform.
If a similar waveform is encoded, then its resulting \ac{hdv} has a high similarity to the first \ac{hdv}.
Each operation is executed on the individual independent components of the \ac{hdv} making them trivial to parallelize.

Traditional \ac{ml} methods such as \ac{dnn} require huge amounts of data and lots of processing power for training \cite{voicehdhyperdimensionalcomputing2017imani}.
\Ac{hdc} promises to reduce these requirements.
Learning from few samples has been demonstrated for the example of seizure detection \cite{oneshotlearningieeg2018burrello}.
The distributed design of \acp{hdv} makes \ac{hdc} very robust against failures in the underlying memory and thus well suited for less reliable low-power emerging memories \cite{inmemoryhyperdimensionalcomputing2020karunaratne}.
The design makes it also robust against noise in the data, e.g., from low-quality aging monitors embedded in the circuit.
Additionally, \ac{hdc} operations are trivial to parallelize to make use of multiple processing units.
All these properties suggest that an ideal aging model can be implemented with \ac{hdc}.

\section{Experimental Setup}
\startcstep
To evaluate the impact of transistor aging on a circuit, the analysis starts at application level.
The activities of the application generates the stimuli for the inputs of the circuit (a NAND gate in this example) as shown in \cref{fig:overview} \cstep{}.
Those stimuli are then propagated to the individual transistors in \cstep{}.
In larger circuits, not every transistor is connected to an input and thus its stimulus depends on the logic inside the circuit. 
Therefore, the circuit has to be simulated to extract the voltage waveforms.
In \cstep{}, the waveforms are provided as an input to the aging models which generate the corresponding degradation trace.
Based on this short-term trace, the \ac{eol} degradation is extrapolated, typically to ten years \cstep{}.
The resulting \ac{eol} \ac{dvt} for each transistor is applied to the circuit \cstep{} and causes an increase of the propagation delay or latency.
Only if this aging-induced shift is considered during design can the system continue functioning properly over its whole lifetime.

This work builds on top of the CARAT framwork \cite{carat2020santen} to simulate circuits with SPICE, extract the voltage waveforms, run the aging models, and simulate again to determine the additional propagation delay. 
A circuit designer can have access to such a framework except for the aging models, which contain sensitive parameters that the foundry does not share. 
Consequently, the whole flow does not benefit the designer because they do not know how much guardband each transistor requires. 
To explore the problem space, the state-of-the-art physics-based \ac{bat} framework \cite{modelingnbtiusing2020mahapatra} is employed.
It estimates the impact of \ac{bti} on different transistor technologies and manufacturing processes.
\Ac{bat} has been validated against several technologies including FinFET, FD-SOI, and nanosheets.
It models the generation of interface and bulk oxide traps as well as hole trapping and other aging effects, including recovery.
The model has been calibrated with experimental measurements to obtain the otherwise confidential parameters.
Such an effort is infeasible for most designers and not possible for technologies in the early prototype stage.
Training data is generated from simple circuits like XOR and NAND.
For all our experiments, the temperature is constant at \SI{90}{\celsius}. 
We discuss other temperature values in \cref{sec:discussion}.
The operating voltage is set to \SI{0.7}{\volt}.

In this work, the traditional \ac{ml}-method \ac{svm} and the emerging brain-inspired \ac{hdc} are investigated.
The training data is presented to both methods as described in \cref{sec:training}.
\Ac{svm} is based on statistical learning frameworks.
Training samples are assigned to one of two groups. To support more classes (i.e., more fine-grained \ac{dvt} values), the problem is mapped to multiple binary classifications. 
The employed Scikit-learn library provides \iac{svm} written in C.
\Iac{svm} can be extended to a nonlinear classifier using the kernel trick.
We perform a grid search to find the best model parameters and utilize the \ac{svm} implementation of the Scikit-learn library \cite{scikit-learn}.
The core parts have been implemented in C.

The recently-proposed OnlineHD is selected as \iac{hdc} implementation \cite{onlinehdrobustefficient2021hernandez-cane}.
It uses the MAP-B \ac{hdv} architecture \cite{multiplicativebindingrepresentation1998gayler}, in which $-1$ and $1$ are the vector components.
The distance between two \acp{hdv} is computed with the cosine similarity.
OnlineHD supports retraining to increase the prediction accuracy.
During retraining, the model is queried with the training dataset and if the prediction is incorrect, the class \ac{hdv} is slightly altered to be more similar to the query \ac{hdv}.
In this work, the number of retraining iterations (epochs) is set to 50 and the learn rate to \num{0.01}.
Similar to \ac{svm}, major parts of OnlineHD have been implemented in C through PyTorch.

In addition, \iac{lstm} model is implemented as an alternative method to the history-based approach with \ac{svm} and \ac{hdc}.
\Ac{lstm} models have been show to work well in sequence to sequence learning applications such as translation tasks \cite{NIPS2014_a14ac55a}.
In this work, an \iac{lstm} encoder-decoder model is trained to predict the full trace based on the input waveform.
The encoder contains two layers of stacked \acp{lstm}, each with 256 units, which learn to map the input waveforms to an internal fixed-size vector representations of size 256.
The decoder is a one layer \ac{lstm} with 256 units.
The decoder is trained to map the fixed internal vector to the \ac{trace}.
Similar to \cite{NIPS2014_a14ac55a}, the performance of the \ac{lstm} model is improved by reversing the input waveforms.

The \ac{lstm} model's performance improved as the number of layers and units in each layer increased, as did the model's complexity.
It was observed that model tends to overfit when the number units is increased above 256.
The \ac{lstm} model's performance tends to deteriorate when the number of segments in the input waveform is greater than 32.

This allows for a fair comparison of the computational demands of both methods and against the physics-based \ac{bat}, all running on an AMD Ryzen 9 3950X . 

\subsection{Datset Generation}
\label{sec:dataset generation}

Circuit designers have access to foundry-provided \acp{pdk} to create and tune their systems.
Typically, the foundry publishes an additional set of \acp{pdk} with aging data under worst-case conditions, which lead to an overestimated guardband.
Actual workloads are far from such worst-case conditions.
Therefore, aging models take the workload into account to predict the expected degradation at \ac{eol} for a single transistor.
The input to the aging model is a \ac{wf} $(V_1, ..., V_{l})$ which is a sequence of $l$ segments where each segment $V_i$ with $i \in \{1, \dots, l\}$ represents the gate voltage applied to the transistor.
The supply voltage can be any of the voltage corners provided by the foundry $V_i \in V_{corners}$.
The time component is included in the \ac{wf} through the segment index, with each segment lasting the same amount of time.

The \ac{wf} is provided to the aging model, which produces a trace $(\Delta\text{\kern-0.15em V}_{th,1}, ..., \Delta\text{\kern-0.15em V}_{th,l})$ reporting a \ac{dvt} for each segment.
The effect of the input voltage is reflected in the output trace $V_i \rightarrow \Delta\text{\kern-0.15em V}_{th,i}$.
However, simply using this mapping as a model does not reflect the voltage dynamics and cannot capture recovery effects.
The $\Delta\text{\kern-0.15em V}_{th,i}$ of segment $V_i$ depends also on the previous segment's $V_{i-1}$, as show in \cref{fig:aging motivation}.

Physics-based models can take the whole \ac{wf} and compute the expected \ac{dvt} for each point in time.
To make such a model accessible to the designer, it has to be replaced with a similarly behaving \ac{ml}-based model to not disclose the confidential technology parameters.
Physics-based models retain the state of the transistor (e.g., the number of defects in the material) during the prediction, which is the basis for their powerful predictive capabilities.
In contrast, lightweight \ac{ml}-based methods do not have such an internal state and have to predict \ac{dvt} iteratively.

\begin{figure}
    \centering
    \includegraphics{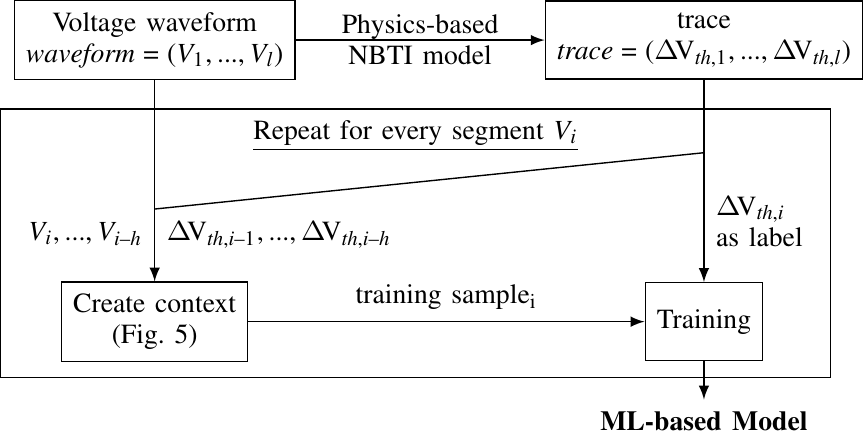}
    \caption{Voltage waveforms derived from circuit-level stimuli are supplied to the physics-based transistor aging model to create training data for the \acl*{ml}-based models. Once they are trained, they take voltage waveforms and predict the degradation trace.}
    \label{fig:flow}
\end{figure}

\subsection{Training data generation}
\label{sec:training}

Training data is generated from 62 standard cells (e.g. XOR, full adder). 
The cells employed in this work have at most five input terminals and no internal state.
With the design of digital circuits in mind, those input terminals are either at \SI{0}{\volt} or at \ac{vdd}.
Random stimuli are applied, which in turn stimulate the internal transistors. 
Through SPICE simulations, the analog waveform for each transistor can be extracted.
The physics-based aging model is then executed to compute the corresponding trace. 
The trace represents the label for a waveform. 
Depending on the type of the cell, each standard cell contains between 4 and 27 pMOS check transistors.
In total, all standard cells contain 414 pMOS transistors.
Thus, 414 waveform-trace pairs, the training samples, can be generated.

While the design of the standard cells is well known, the designer's circuits is their intellectual property that cannot be shared with third parties like the foundry. 
Therefore, we mimic the application scenario for a circuit designer and generate the test set from transistors in larger circuits. 
In this work, two circuits are explored. 
First, an adder for two 8-bit numbers with 111 transistors. Second, a 32-bit MAC unit, that multiplies an 8-bit weight with an 8-bit input and accumulates the result with a 32-bit partial sum.
The circuit contains 1395 pMOS transistors. 
The inputs of each circuit are stimulated with random data for an unbiased evaluation. 
A circuit designer would simulate their typical workload patterns.
Similar to the standard cells, the inputs propagate through the circuit and waveforms for each transistor are extracted. 
In other words, the designer extracts waveforms representing their workload.  
For evaluation purposes, the physics-based aging model is employed again to compute the traces as a ground truth. 
The number of consecutive addition or MAC operations can be set to generate waveforms of various lengths.
The longer the trace, the more it challenges the ML model since more features (input voltages) have to be considered.

\begin{figure}
    \centering
    \includegraphics{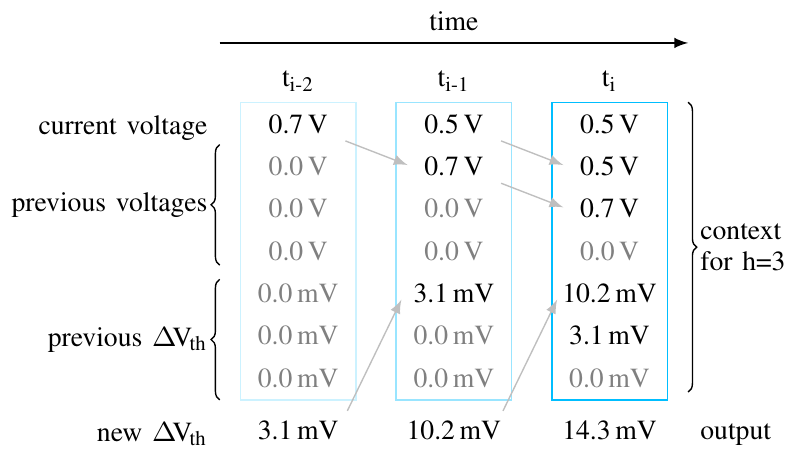}
    \caption{Some history is added to the current input voltage to better capture the voltage dynamics. In this example $h=3$, i.e., the input voltage and $\Delta\text{\kern-0.15em V}_{th,i}$ from t\textsubscript{i-1}, t\textsubscript{i-2}, and t\textsubscript{i-3} are included.}
    \label{fig:history}
\end{figure}

\section{Scenario 1: Predicting a Full Trace}
\label{sec:scenario full trace}
The objective is to predict a \ac{dvt} for each segment of the \ac{wf}.
In contrast to \iac{lstm}, \iac{svm} or \iac{hdc} cannot directly convert a sequence to another. 
Hence, the \acp{wf} have to be processed to make them learnable by the latter models.
The training procedure for one \ac{wf} is sketched in \cref{fig:flow}.
Since the current state of the transistor is not available for training, the voltage dynamics have to be captured with a history of $h$ previous \ac{wf} segments.
However, such a snippet of the \ac{wf} sequence is not bound to a specific point in time or, more importantly, to the current internal state of the transistor.
Setting $h = l$ (i.e., include all segments) is not viable due to the prohibitively large parameter space.
Thus, a history of the $h$ previous \ac{dvt} values is included as well. 
The combination of voltage and \ac{dvt} provides a more detailed context for training.
\Cref{fig:history} visualizes the information contained in three training samples for $h=3$.
The label for each sample at time $i$ is the $\Delta\text{\kern-0.15em V}_{th,i}$ of the segment taken from the \ac{trace}.
The $\Delta\text{\kern-0.15em V}_{th,i}$ is quantized to discrete labels for classification.

The results show that the \ac{svm} and \ac{hdc} models have a bias in their predictions. 
Although their predictions follow the traces in general, the nominal \ac{dvt} values often deviate.
A multiplier can reduced this offset.
After the model training is complete, it is used on the training set itself to predict the \acp{trace}.
The disagreement between the \ac{ml}-based and the physics-based model is analyzed and the resulting average deviation is used as a multiplier during inference.

\subsection{Inference}
\label{sec:inference}

During inference, the same data representation, described above, is used for \ac{svm} and \ac{hdc}.
This representation includes the $h$ previous \ac{dvt}.
However, only the \ac{wf} is available during inference.
Hence, the \ac{dvt} values have to be predicted online during inference.
They are then adjusted with the multiplier to be directly used to predict the next segment.
For the first segment $i=1$, the initial \ac{dvt} and the ``previous'' \ac{dvt} is set to \SI{0}{\milli\volt}, as shown in \cref{fig:history}.
In effect, only the input voltage $V_1$ determines $\Delta\text{\kern-0.15em V}_{th,1}$.
The predicted $\Delta\text{\kern-0.15em V}_{th,1}$ is then used to create the context for segment $i=2$, $\Delta\text{\kern-0.15em V}_{th,1}$ and $\Delta\text{\kern-0.15em V}_{th,2}$ for segment $i=3$, and so forth.
Due to this recursive process, prediction errors multiply requiring high precision.

\subsection{Evaluation}

\begin{figure}
    \centering
    \includegraphics{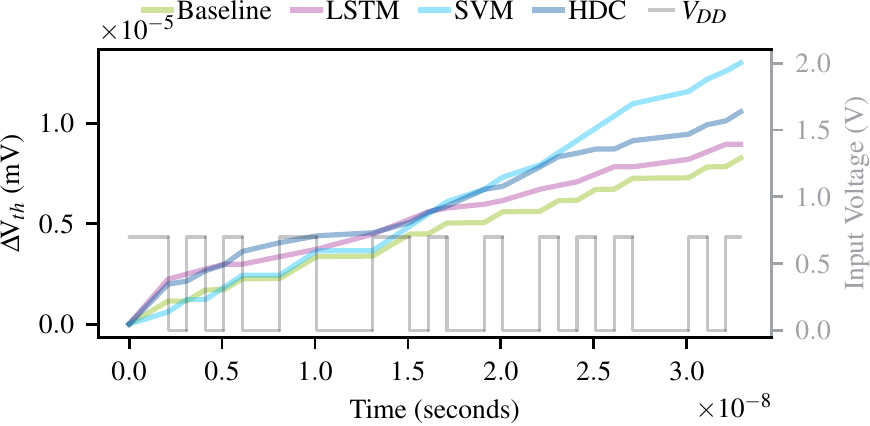}
    \caption{Example of a waveform (gray), the baseline trace from the physics-based model (green) and the predicted traces from various \ac{ml} models.}
    \label{fig:traces}
\end{figure}

\begin{figure*}
   \begin{minipage}{0.66\textwidth}
     \centering
     \includegraphics{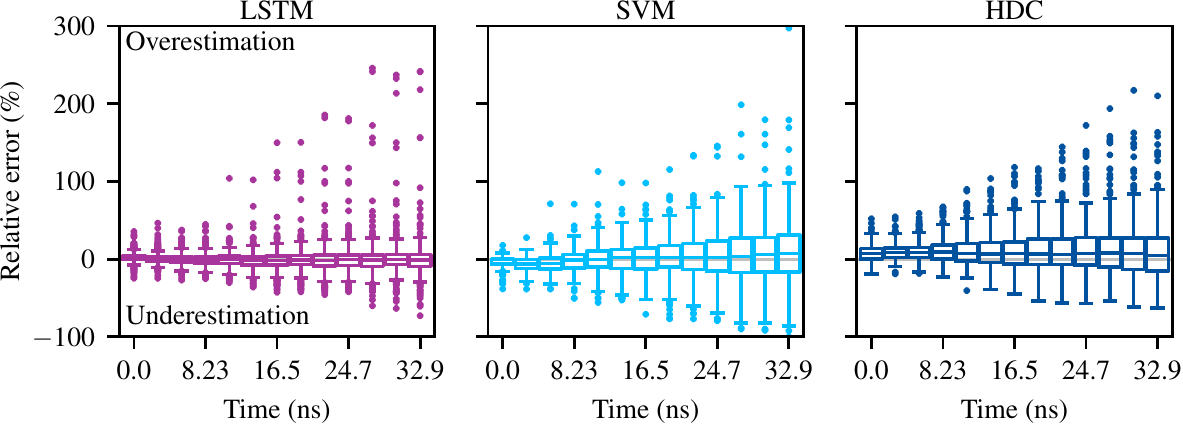}
     \caption{The \ac*{svm} and \ac{hdc} models rely on their own previous outputs for the next prediction. Hence, the error accumulates, which is represented by the increase in the relative error. The \ac{lstm} directly translates the whole sequence and achieves a higher accuracy although outliers are as bad as in other models. Training and test are preformed on the adder circuit.}\label{fig:example}
   \end{minipage}\hfill
   \begin{minipage}{0.32\textwidth}
     \centering
     \includegraphics{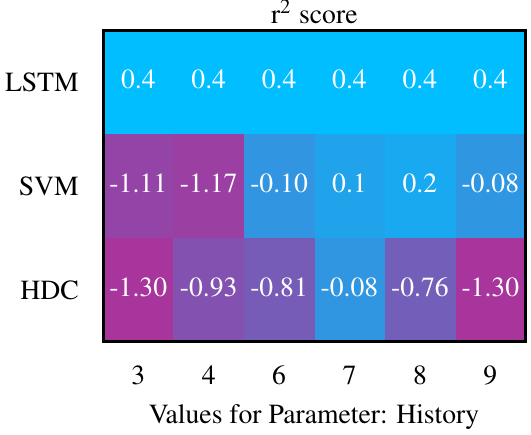}
     \caption{Mean r\textsuperscript{2} scores for different histories $h$ with \iacs*{hdc} dimension of 20k for the adder dataset.}\label{fig:dimension and history}
   \end{minipage}
\end{figure*}

The performance of the \ac{ml}-based models under a variety of different aspects is evaluated.
The datasets are generated and split into training and testing set with a \SI{70}{\percent} split.
As a metric, the \acl{re} per segment $RE_i = (ML_i - BAT_i) / BAT_l \, * \SI{100}{\percent}$ is used.
$ML_i$ and $BAT_i$ are the predictions for segment $i$ from the \ac{ml}-based and physics-based models, respectively.
The difference is divided by $BAT_l$, the final \ac{dvt}.
Overestimating the degradation results in a positive \ac{re}, underestimating it in  a negative.
The results in \cref{fig:example} show balanced models with a tendency for overestimation.

\subsection{Dimension of the HDC Model}
The dimension of the \acp{hdv} determines their capacity to store information.
The higher the dimension, the higher the expected accuracy.
This increase levels off at an application-specific point, which is not known a priori.
A higher dimension also correlates with more costly operations and higher memory requirements.
Both costs are not the primary concern during design time.
Therefore, \ac{hdc}-based models with high dimensions above \num{10000} are feasible.

In this work, dimensions from \num{1000} to \num{20000} vector elements are explored.
Contrary to the initial assumption, a higher dimension does necessarily not result in higher accuracy.
The mean \ac{re} for different dimensions is shown in \cref{fig:dimension and history}.
The highest dimension of \num{20000} performs best on average over different $h$, but even the lowest dimension with \num{1000} outperforms others.
These results indicate that the \ac{hdc} model has unused capacity available.

\subsection{Impact of History h}
The parameter $h$ determines how many previous segments are taken into account to predict the next segment.
The \ac{svm} performs best with $h=8$.
For \ac{hdc}, the combination with the dimension has to be considered.
More history requires a higher capacity of the model to contain the information.
While this capacity is available with the high dimensions, the results suggest an oversaturation of the query \ac{hdv} with the same voltage \ac{hdv}.
A different encoding is expected to mitigate this issue.
The overall best performances for \ac{hdc} are achieved with $h$ around seven.

The hyperparameters dimension and $h$ can be selected based on the model's performance on the training data.
Our analysis of circuits, discussed in \cref{sec:scenario eol}, shows that different settings are required depending on the workload characteristics.
The best model is selected by the foundry and send to the designer.

\subsection{Reduction of Model Execution Time}
In the \ac{hdc} model, the complex differential equations of the physics-based model are replaced with simple operations on integer vectors. 
The performance advantages are reflected by a reduced execution time shown in \cref{tab:times}.
Predicting a 32-segment trace for the 8-bit adder takes \SIrange{29}{88}{\milli\second} for a dimension of \num{1000} and \num{20000}, respectively.
This is up to 30X faster compared to the physics-based model with \SI{602}{\milli\second} or the \ac{svm} with \SI{624}{\milli\second}.
The time for training varies with $h$, but it is consistently lower for the \ac{hdc} model compared to the \ac{svm}.
OnlineHD utilizes multiple CPU cores to reduce the training time.
The \ac{lstm} takes the most time, even longer than the physics-based model, but achieves the best accuracy. 

\begin{table}
	\caption{Execution times for the 8-bit adder circuit.}
	\label{tab:times}
	\vspace{0.5ex}
	\centering
	\footnotesize
	\begin{tabular}{lcc}
		\toprule
		Task                           &         &                     Wall-clock time                      \\ \midrule
		Training set generation        & (total) &                   \SI{409.0}{\second}                    \\
		\acs*{hdc} training            & (total) &   \SIrange[range-phrase={~--~}]{34.2}{152.3}{\second}    \\
		\acs*{svm} training            & (total) &                   \SI{407.7}{\second}                    \\ \midrule
		Physics-based trace prediction & (mean)  &                \SI{602.2}{\milli\second}                 \\
		\acs*{hdc} trace prediction    & (mean)  & \SIrange[range-phrase={~--~}]{28.7}{88.1}{\milli\second} \\
		\acs*{svm} trace prediction    & (mean)  &                \SI{624.1}{\milli\second}                 \\
		\acs*{lstm} trace prediction   & (mean)  &                \SI{1006.7}{\milli\second}                 \\ \bottomrule
	\end{tabular}
\end{table}

\section{Scenario 2: End-of-lifetime Aging}
\label{sec:scenario eol}
Recreating the degradation trace is useful in evaluating short-term aging effects \cite{van2016designing}.
To predict the degradation at the \ac{eol} of the device, and thus for circuit designers to add sufficient guardbands, the whole trace is not necessary. 
The extrapolation model for \ac{bti} considers the \ac{wf} as well as the last \ac{dvt} value.
Hence, an \ac{ml} model is sufficient for \ac{eol} \ac{dvt} estimation if it can predict this last value.
Consequently, the challenge transforms from a recursive trace reconstruction to a simpler regression.
With the focus on long-term aging, the final impact of inaccurate predictions from \ac{ml} models can be evaluated at circuit level.
The physics-based \ac{bat} is replaced with an \ac{ml} model to provide the short-term aging value. 
This result is then processed further by the CARAT framework to predict the aging-induced shift in the circuit's propagation delay. 

\subsection{Model Training and Evaluation}

Many \ac{ml} algorithms exist to solve regression problems.
The input is \iac{wf}, where each segment acts as a feature and the predicted output is the last \ac{dvt} value. 

\Iac{svm} can also be used for regression and is then referred to as an \textbf{SVR}.
The implementation is based on the Scikit-learn library \cite{scikit-learn} and a grid search is done for hyperparameter tuning.
The SVR has an Radial Basis Function kernel, a gamma value of 0.001, and a C of 100.
An \textbf{\ac{mlp}} is implemented with PyTorch \cite{NEURIPS2019_bdbca288}. 
It has a total of three layers with 128 neurons in the hidden layer.
The output layer is a single neuron.
In contrast to classification, this single neuron returns a floating point value representing the last \ac{dvt}.
An \textbf{\ac{hdc}} classifier can be used for regression by quantizing the \ac{dvt} values and treating those as classes.
For comparison, a \textbf{worst-case model} is created.
With \ac{bti}, the pMOS transistor ages if no gate voltage is applied.
Hence, the worst case assumes that the transistor is turned off and only turns on at the end of the simulated time frame to maximize the aging effect.

Each model is trained and evaluated on three circuits.
The dataset generation is described in detail in \cref{sec:dataset generation}.
The aging extrapolation models for \ac{bti} depend on the last \ac{dvt} value and the \ac{wf}.
But instead of the physics-based aging model, the \ac{ml} models are employed.
The predictions are compared with the output of the physics-based model as a baseline.
As an accuracy metric, r\textsuperscript{2} score is select, with a value of `1' as a perfect match. 

\subsection{Results at Circuit Level}

\begin{figure*}
    \centering
    \includegraphics{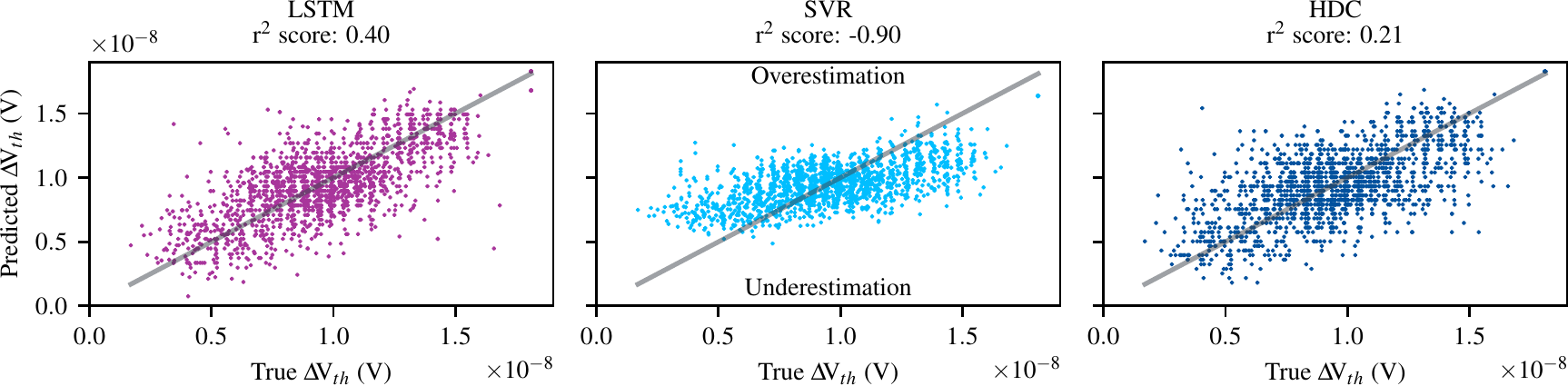}
    \caption{Training and testing on the same circuit provides a baseline for the complexity of the problem to predict the final \ac{dvt} value based on the trace. The \ac{lstm} predicts the whole trace but only the last value is considered in the evaluation.}
    \label{fig:dvt adder}
\end{figure*}

\begin{figure*}
    \centering
    \includegraphics{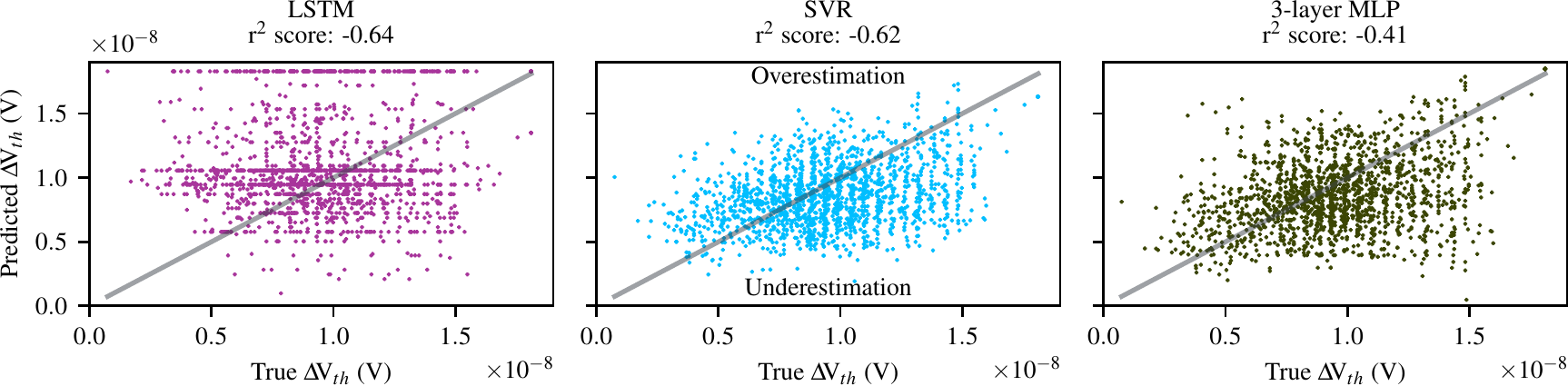}
    \caption{Standard cells are provided by the foundry and the base for many circuit designs. However, the training dataset generated from them is too small for the \ac{ml} methods to sufficiently learn and generalize. Hence, the inference results with the adder circuit are worse compared to \cref{fig:dvt adder}.}
    \label{fig:dvt std add}
\end{figure*}

\begin{figure*}
    \centering
    \includegraphics{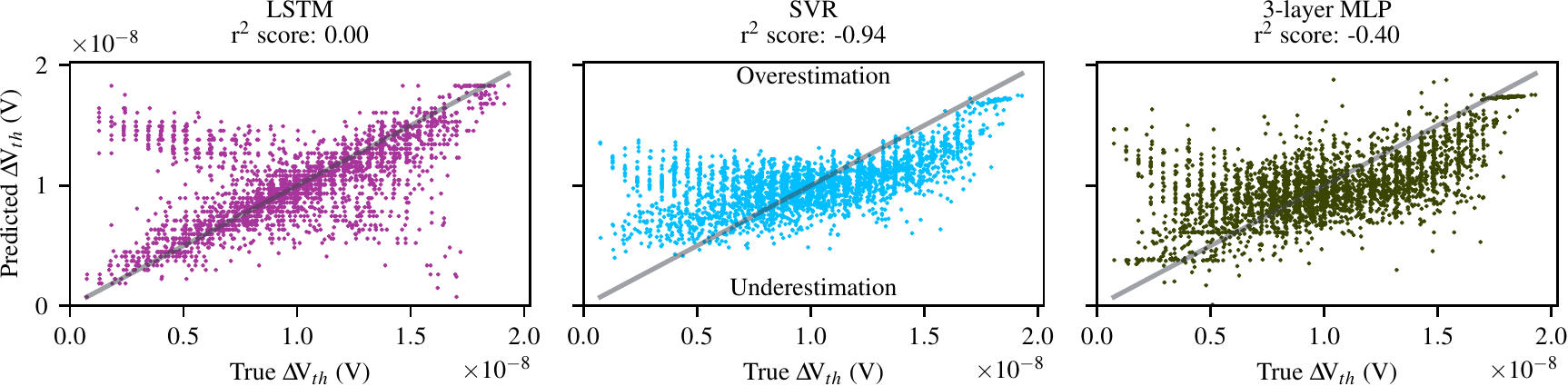}
    \caption{The models are trained on the adder circuit and used to predict the degradation in the MAC circuit. The large dataset from the adder allows the \ac{lstm} to train and provide adequate results. The outliers are analyzed in \cref{sec:outlier analysis}. }
    \label{fig:dvt add mac}
\end{figure*} 
To judge the complex of the problem, the models are trained and tested on the adder circuit.
Three sets of random inputs are generated for training, hyperparameter tuning, and testing.
The results are presented in \cref{fig:dvt adder} and show the correlation between the baseline physics-based last \ac{dvt} values and the ML-based ones for all transistors.
The r\textsuperscript{2} scores are given above the plots and show that the best \ac{ml} approach is the \ac{lstm} model.
An r\textsuperscript{2} score of 0.37 was achieved with a training for 500 epochs, two hidden layers with 25 units per layer, and the L1 loss function.
Although there is some spread around the baseline, the model's ability to predict the \ac{dvt} is clear.
A similar picture is given by the \ac{mlp}, the predictions are correlated with the baseline values. 
For \ac{hdc}, the spread is even larger but still follows the baseline.
The model is trained with a dimension of 4000 for 50 epochs.

While \ac{hdc} has the highest spread, the mean aging-induced shift in the propagation delay at circuit level is equal to the baseline. 
\Cref{tab:res delay adder adder} compares the different models and also includes the worst-case model with constant aging stress. 
Both, \ac{hdc} and \ac{mlp}, overestimate the impact overestimate aging, which is preferable to the \ac{lstm}, which underestimates the impact and thus could lead to insufficiently small guardbands.
However, even with doubling the \ac{ml}-based predictions to save guard against underestimation, the \ac{ml}-models still outperform the worst-case model by a factor of three. 

Training and predicting for the same circuit would require that the circuit designers share details with the foundry, which would train the model.
To minimize data sharing, the foundry can train a model on their standard cells and provide those models to designers. 
However, the results plotted in \cref{fig:dvt std add} show a significant degradation of the quality of the predictions.
The r\textsuperscript{2} scores drop below zero and the models struggle to generalize. 
Worse, the \ac{lstm} and the \ac{mlp} predict low \ac{dvt} values although the baseline values are close to the maximum (prediction in lower right corner). 
While the spread of the \ac{svm} has increased compared to a training with the adder, the maximum prediction errors are smaller than with other \ac{ml} models. 
The \ac{hdc} model has failed to generalize and is not included in the results. 

Similar and better results are shown in \cref{fig:dvt add mac} for training on the adder and testing on the MAC circuit.
First, the \ac{lstm} has sufficient data to train and can predict most samples with a low error. 
Nevertheless, outliers can cause incorrect guardband estimations. 
Second, while the \ac{svm}'s r\textsuperscript{2} is the lowest, it underestimations the least preventing severely incorrect guardband estimations. Overestimations are limited to smaller \ac{dvt} values and in total the \ac{svm} model achieves a mean relative error of \SI{1.7}{\percent} compared to the \ac{lstm}'s \SI{3}{\percent}.
Finally, the largest \ac{dvt} value in the MAC dataset is higher than in the adder and this behavior is not not captured by the \ac{ml} models, they are limited by their training. 
This is evident by the horizontal cluster in the top right.

\subsection{Error Analysis}
\label{sec:outlier analysis}

\begin{figure}
    \centering
    \includegraphics{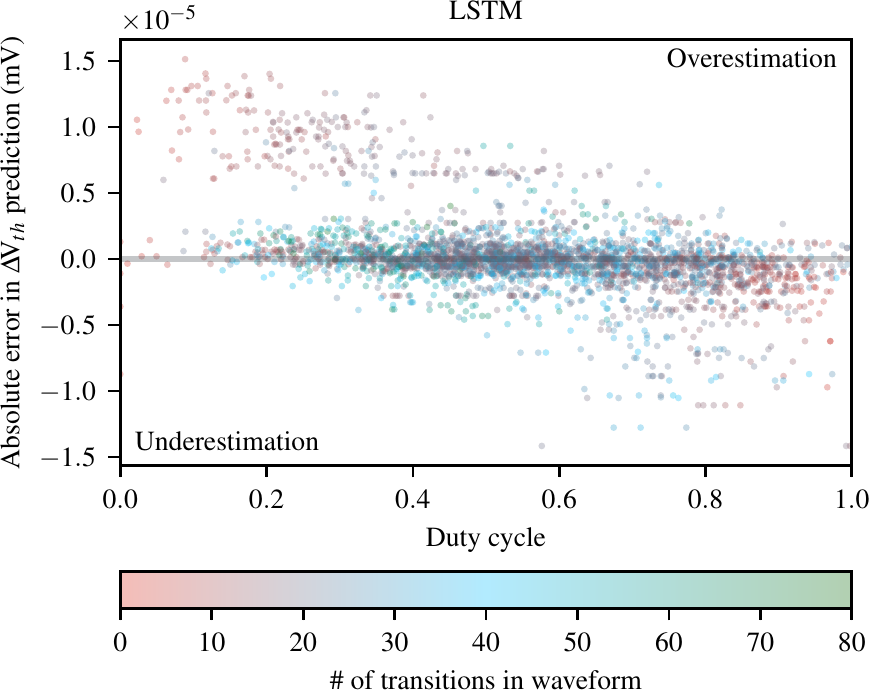}
    \caption{Analysis of the \acs*{lstm} model for the MAC circuit. The largest prediction errors are correlated with a low duty cycle and a low number of voltage transitions.}
    \label{fig:error analysis}
\end{figure}

While many predictions of the \ac{lstm} are within a tolerable error range, there are outliers that are either under- or overestimated as shown in \cref{fig:dvt add mac}.
The same samples are plotted in \cref{fig:error analysis} for an error analysis. 
First, overestimated samples have a lower duty cycle and especially fewer voltage transitions in the waveform.
In other words, transistors that are off most of the time and change their on/off state seldom. 
Overestimations have a negative impact on the circuit's timing because guardband are designed unnecessary large. 
However, they do not lead to failure of the device, in contrast to underestimations.
The impact of aging is underestimated for some transistors with a duty cycle above \num{0.6}.
Their waveforms have an average amount of transitions.
This combination of duty cycle and number of transitions is not a defining feature for underestimations by the \ac{lstm} model.
Hence, it is impossible to derive a simple rule-based solution to contain the potential timing errors due to insufficient guardbands.

The SVR shows a similar pattern. 
Overestimations correlate with a low duty cycle combined with a low number of transitions in the waveform. 
Underestimations are not as frequent and as pronounced.
While they occur mainly above a duty cycle of \num{0.4}, worst-case underestimations do not correlate with the number of transitions in the waveform.
Hence, similar to the \ac{lstm}, a simple rule-based error reduction cannot be derived. 
In summary, the models perform well for most samples but outliers, especially underestimations, still pose a challenge. 

\begin{table}[t]
    \centering
    \caption{Aging-induced delay for train and test on adder circuit.}
    \label{tab:res delay adder adder}
    \begin{tabular}{lSSSSS}
    	\toprule
    	Delay (\si{\pico\second}) & {Baseline} & {LSTM} & {SVR}   & {HDC} & {Worst Case} \\ \midrule
    	min                       & -2.08      & -2.08  & -0.40   & -1.81 & -2.12        \\
    	mean                      & 4.88       &  4.66  &  5.33   &  4.88 & 31.36        \\
    	max                       & 12.60      & 11.70  & 11.70    & 13.90 & 61.10        \\ \bottomrule
    \end{tabular}
\end{table}

\begin{table}[t]
    \centering
    \caption{Aging-induced delay for train on standard cells and test on adder circuit.}
    \begin{tabular}{lSSSSS}
    	\toprule
        Delay (\si{\pico\second}) & {Baseline} & {LSTM} & {SVR} & {MLP} & {Worst Case} \\ \midrule
        min                       & -2.08      & -4.33  & -0.93 & -1.14 &  -2.13       \\
        mean                      &  4.76      &  6.76  &  5.06 &  5.19 &  31.52       \\
        max                       & 10.90      & 18.80  & 12.60 & 15.10 &  61.10       \\ \bottomrule
    \end{tabular}
    \label{tab:res delay std adder}
\end{table}

\begin{table}[t]
    \centering
    \caption{Aging-induced delay for train on adder and test on MAC circuit.}
    \begin{tabular}{lSSSSS}
    	\toprule
        Delay (\si{\pico\second}) & {Baseline} & {LSTM} & {SVR} & {MLP} & {Worst Case} \\ \midrule
        min                       & -1.80      & -2.90  & -2.10 & -3.80 &  -70.70       \\
        mean                      &  5.03      &  5.58  &  5.31 &  4.94 &   88.67       \\ 
        max                       & 15.00      & 14.90  & 12.60 & 11.50 &  450.91       \\\bottomrule
    \end{tabular}
    \label{tab:res delay adder mac}
\end{table}
 
\section{Discussion}
\label{sec:discussion}
The focus of this work is on \ac{bti}, the dominant degradation effect in current transistor technology \cite{modelingnbtiusing2020mahapatra}.
Nevertheless, PBTI and \ac{hcd} also play an important role. 
Their impact on the transistor has to be considered as well to design the circuit with small yet sufficient timing guard bands. 
Hence, an investigation into replacing those models with \ac{ml}-based models is necessary.
Preliminary results suggest that the methods explored in this work are challenged by the different types of stimuli driving those degradation effects. 
In \ac{bti}, the on/off time is the dominant factor whereas in \ac{hcd} the number of transitions has to be considered, among other stimuli. 

Aging effects also depend heavily on the temperature of the transistor. 
The experiments in this work assume a constant temperature of \SI{90}{\celsius}. 
However, the temperature of a transistor keeps changing between high-load phases and standby states of the overall system the circuit is integrated
in. 
Those dynamic changes have to be investigated and included for a temperature-aware \ac{ml}-based trace prediction.

\section{Conclusion}
Accurate physics-based aging models include confidential technology and material parameters.
Thus, such models are not available to circuit designers to optimize their designs under the actual impact of aging.
This work explores the applicability of \ac{ml}-based methods to train on the physics-based models, in particular traditional \ac{svm}, \ac{lstm}, and brain-inspired \ac{hdc}.
While \ac{ml}-based models can predict the impact of aging for most transistors accurately, outliers can be over- or underestimated.
Nevertheless, the explored \ac{ml}-based methods predict the degradation about 3x more precise than available worst-case models. 
For the first time, circuit designers have access to an accurate aging model which is indispensable for efficient designs.
This work opens the door to narrow the boundary between foundry and circuit designers.

\section*{Acknowledgment}
The authors would like to thank Victor van Santen for his support with the physics-based aging models. 
This research was partially supported by Advantest as part of the Graduate School ``Intelligent Methods for Test and Reliability'' (GS-IMTR) at the University of Stuttgart.

\renewcommand*{\bibfont}{\footnotesize}
\renewcommand*{\UrlFont}{\rmfamily}

\balance
\printbibliography

\end{document}